\documentclass[letterpaper, 10 pt, conference]{ieeeconf}

\IEEEoverridecommandlockouts

\usepackage{cite}
\usepackage{ulem}
\usepackage{acronym}
\usepackage{caption}
\usepackage{leftidx}
\usepackage{listings}
\usepackage{graphicx}
\usepackage{textcomp}
\usepackage{multirow}
\usepackage{booktabs}
\usepackage{colortbl}
\usepackage{subcaption}
\usepackage{algorithmic}
\usepackage{amsmath,amssymb,amsfonts}

\newcommand{\norm}[1]{\left\lVert#1\right\rVert}
\newcommand{\abs}[1]{\left\lvert#1\right\rvert}
\usepackage{booktabs}
\usepackage{array}
\usepackage{savesym}
\savesymbol{checkmark}
\usepackage{dingbat}
\usepackage{dingbat}
\usepackage{bbding}
\usepackage{colortbl}
\usepackage[table]{xcolor}

\usepackage[utf8]{inputenc}
\usepackage{pgfplots}
\DeclareUnicodeCharacter{2212}{−}
\usepgfplotslibrary{groupplots,dateplot}
\usetikzlibrary{patterns,shapes.arrows}
\pgfplotsset{compat=newest}

\usepackage{hyperref}
\usepackage{cleveref}
\DeclareMathOperator*{\argmax}{argmax} 

\acrodef{TP}{True Positive}
\acrodef{FP}{False Positive}
\acrodef{FN}{False Negative}
\acrodef{VSLAM}{Visual SLAM}
\acrodef{STD}{Standard Deviation}
\acrodef{ROS}{Robot Operating System}
\acrodef{RMSE}{Root Mean Square Error}
\acrodef{ATE}{Absolute Trajectory Error}
\acrodef{RANSAC}{RANdom SAmple Consensus}
\acrodef{CNN}{Convolutional Neural Network}
\acrodef{LiDAR}{Light Detection And Ranging}
\acrodef{SLAM}{Simultaneous Localization and Mapping}

\definecolor{red}{HTML}{fd8f8f}
\definecolor{greend}{HTML}{57e377}
\definecolor{greenl}{HTML}{b8fb8a}
\definecolor{yellow}{HTML}{fefdb4}
\definecolor{orange}{HTML}{ffd5ab}

\colorlet{red}{red!50}
\colorlet{yellow}{yellow!50}
\colorlet{greenl}{greenl!50}
\colorlet{greend}{greend!50}
\colorlet{orange}{orange!50}

\title{\LARGE \bf Category-level Meta-learned NeRF Priors for Efficient Object Mapping}

\author{
    Saad Ejaz$^{1}$, Hriday Bavle$^{1}$, Laura Ribeiro$^{1}$, Holger Voos$^{1}$, and Jose Luis Sanchez-Lopez$^{1}$ 
    \thanks{$^{1}$Authors are with the Automation and Robotics Research Group, Interdisciplinary Centre for Security, Reliability, and Trust (SnT), University of Luxembourg, Luxembourg. Holger Voos is also associated with the Faculty of Science, Technology, and Medicine, University of Luxembourg, Luxembourg. \tt{\small{\{saad.ejaz, hriday.bavle, laura.ribeiro, holger.voos, joseluis.sanchezlopez\}}@uni.lu}}
    \thanks{This research was funded, in whole or in part, by the Luxembourg National Research Fund (FNR) under the DEUS Project (Ref. C22/IS/17387634/DEUS) and the MR-Cobot Project (Ref. 18883697/MR-Cobot).}
    \thanks{*For the purpose of open access, and in fulfillment of the obligations arising from the grant agreement, the author has applied a Creative Commons Attribution 4.0 International (CC BY 4.0) license to any  Author Accepted Manuscript version arising from this submission.}
}
\begin{document}

\maketitle
\thispagestyle{empty}
\pagestyle{empty}

\begin{abstract}
In 3D object mapping, category-level priors enable efficient object reconstruction and canonical pose estimation, requiring only a single prior per semantic category (e.g., chair, book, laptop, etc.). 
DeepSDF has been used predominantly as a category-level shape prior, but it struggles to reconstruct sharp geometry and is computationally expensive. 
In contrast, NeRFs capture fine details but have yet to be effectively integrated with category-level priors in a real-time multi-object mapping framework. 
To bridge this gap, we introduce PRENOM, a Prior-based Efficient Neural Object Mapper that integrates category-level priors with object-level NeRFs to enhance reconstruction efficiency and enable canonical object pose estimation. 
PRENOM gets to know objects on a first-name basis by meta-learning on synthetic reconstruction tasks generated from open-source shape datasets. 
To account for object category variations, it employs a multi-objective genetic algorithm to optimize the NeRF architecture for each category, balancing reconstruction quality and training time. 
Additionally, prior-based probabilistic ray sampling directs sampling toward expected object regions, accelerating convergence and improving reconstruction quality under constrained resources.
Experimental results highlight the ability of PRENOM to achieve high-quality reconstructions while maintaining computational feasibility.
Specifically, comparisons with prior-free NeRF-based approaches on a synthetic dataset show a 21\% lower Chamfer distance.
Furthermore, evaluations against other approaches using shape priors on a noisy real-world dataset indicate a 13\% improvement averaged across all reconstruction metrics, and comparable pose and size estimation accuracy, while being trained for 5$\times$ less time.

Code available at: \hyperlink{https://github.com/snt-arg/PRENOMv1}{https://github.com/snt-arg/PRENOM}

\end{abstract}
\section{Introduction}
\label{sec_intro}

Object-level mapping has gained significant attention for its ability to separate objects from cluttered environments for high-level robotic tasks. Recent works use compact primitives such as cuboids \cite{wu2023object} and quadrics \cite{han2023sq} to approximate object bounds, while reconstructing dense geometry using Signed Distance Fields (SDFs) \cite{wu2022object, wu2023objectsdf++}, Neural Radiance Fields (NeRFs) \cite{kong2023vmap, han2023ro}, and Gaussian splats \cite{yang2024gaussianobject}. Typical methods operate without explicit priors and treat all objects uniformly, overlooking object semantics. In contrast, 3D priors capture common geometric properties, providing valuable cues for reconstruction while also enabling canonical pose estimation of objects.
\begin{figure}[t]
    \centering
    \includegraphics[width=1.0\columnwidth]{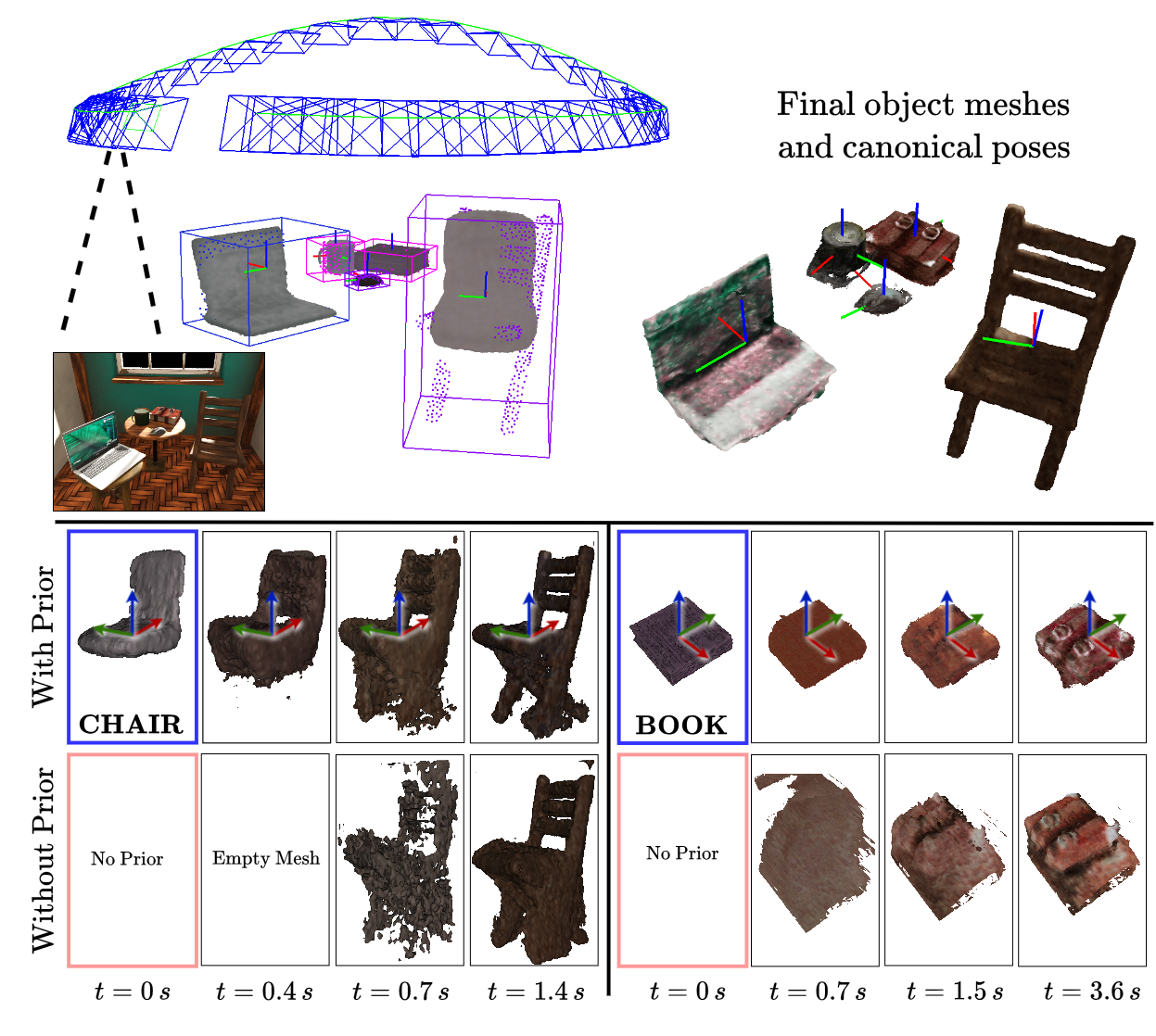}
    \caption{PRENOM uses category-level priors to accelerate object-level NeRF training and localize objects in a canonical frame. Top left shows priors loaded during mapping (displayed all at once for illustration). PRENOM operates online, training objects in parallel while initializing them sequentially as they are observed. The bottom rows demonstrates how priors provide a reasonable initial estimate, enabling faster and more accurate reconstruction. Time indicates dedicated GPU training time.}
    \label{fig:teaser}
\end{figure}

These priors can be broadly classified into three categories. Instance-level priors \cite{salas2013slam++, merrill2022instance} offer high accuracy, but require knowledge of the environment, through precise CAD models of each object instance, limiting their applicability to specialized use cases. On the other end, generalized priors \cite{GOM} use a single prior for all objects but rely on bulky diffusion models, making them computationally intensive even for a single object.
Category-level priors are a suitable trade-off as they rely solely on semantic labels, and thus can work with unseen objects within predefined categories of interest. Recent works represent such priors using variational autoencoders \cite{sucar2020nodeslam},  shape embeddings \cite{li2020frodo}, and DeepSDFs \cite{wang2021dsp, zou2022objectfusion, liao2024uncertainshapepose}. While these methods offer reasonable global shape and pose estimation, they are computationally demanding, requiring high-end GPUs; use identical prior designs for all categories, ignoring differences in shape complexity; and in most scenarios struggle with reconstructing sharp geometry. 


To address these limitations, we present PRENOM, a Prior-based \textit{Efficient} Neural Object Mapper that represents objects by lightweight NeRF models, with meta-learned \cite{tancik2021learned} parameter initializations as category-level priors, within a real-time RGB-D multi-object mapping framework. 
Having been trained on large shape datasets, these priors can localize objects within a canonical frame while also benefiting from the high-fidelity reconstructions typically associated with NeRFs.
Further integrating them with neural architectures optimized per-category and prior-based probabilistic ray sampling, we achieve faster training convergence (see \Cref{fig:teaser}), enabling accurate reconstructions with limited GPU resources.




The primary contributions of this paper are as follows:
\begin{itemize} 
\item An online object mapping framework that uses lightweight object-level NeRFs with category-level meta-learned priors to enable accurate object reconstruction on resource-constrained systems. 
\item A pipeline for optimizing NeRF architectures per object category to achieve a balance between reconstruction performance and training speed. 
\item Integration of prior-based probabilistic ray sampling in object-level NeRFs for improved convergence leading to better reconstruction quality.
\end{itemize}
\section{Related Works}
\label{sec_related}

\subsection{3D Object Mapping}
Object-level mapping has been extensively explored in the context of SLAM, where camera poses and object states are jointly estimated by enforcing multi-view consistency \cite{wu2023object, han2023sq, wang2021dsp, sucar2020nodeslam}. Alternatively, it has also been studied independently, taking advantage of known camera poses to focus purely on object representation and reconstruction \cite{li2020frodo, wu2022object, abou2022implicit, kong2023vmap, han2023ro, liao2024uncertainshapepose, GOM}. While the former approach prioritizes simple, optimizable object representations to support joint estimation, the latter decouples localization and mapping, shifting the focus on improving reconstruction fidelity. To achieve this, implicit neural object representations, such as object-level SDFs \cite{wu2022object, wu2023objectsdf++} and object-level NeRFs \cite{yang2021objNerf, abou2022implicit, kong2023vmap, han2023ro}, have been widely adopted to generate high-quality meshes. Our approach falls in this latter category.

Considering only object representation, RO-MAP \cite{han2023ro} and \cite{abou2022implicit} are the closest to our approach; however, their reliance on random weight initialization and uniform ray sampling results in slow training, requiring a high-end GPU for real-time multi-object mapping. 
Moreover, scale ambiguity in the monocular RO-MAP results in noisy object pose estimation, which is why we opt for an RGB-D pipeline.

\subsection{Category-level Priors for Object Mapping}
NodeSLAM \cite{sucar2020nodeslam} employs variational autoencoders as category-level shape priors to predict occupancy grids, jointly estimate the shape and pose of objects. However, it struggles with reconstructing complex shapes.
Frodo \cite{li2020frodo} represents shape priors using encoder-decoder networks to model objects as point clouds and SDFs. Unlike our sequential method, it operates in batch mode.

Several works \cite{wang2021dsp, zou2022objectfusion, liao2024uncertainshapepose} adopt the DeepSDF architecture \cite{DeepSDF} as a shape prior, learning latent embeddings as shape priors, and training decoder networks to predict object SDFs. 
These approaches require long training times of several seconds per object on a low-end GPU while having limited capability to reconstruct fine geometry. 
Moreover, these methods also jointly estimating the object pose with shape reconstruction. While this can refine object localization, it increases training time and makes reconstruction accuracy dependent on state estimation quality. In contrast, our approach follows \cite{sucar2020nodeslam} by estimating the object pose, before initializing reconstruction. However, unlike the CNNs used by \cite{sucar2020nodeslam}, 
we rely solely on our priors to estimate and refine the pose.



\section{Methodology}
\label{sec_proposed}

\begin{figure*}[ht]
    \vspace{5pt}
    \centering
    \includegraphics[width=1.0\textwidth]{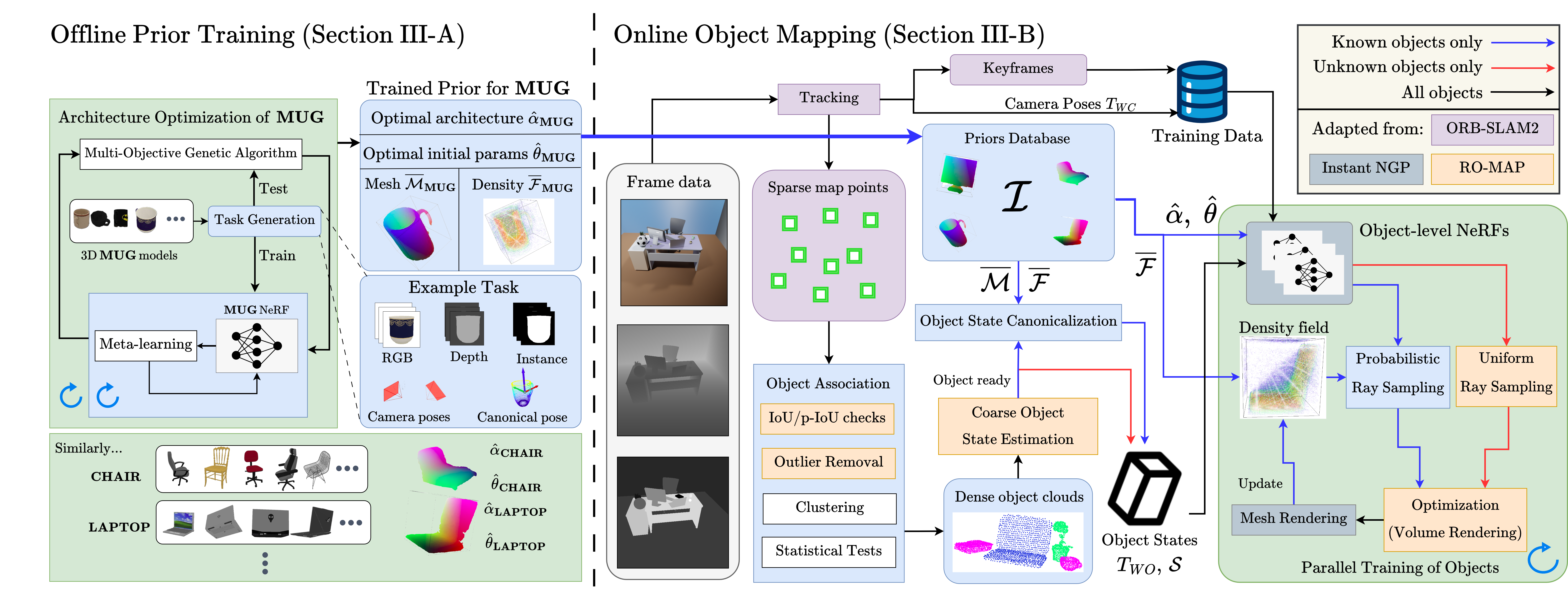}
    \caption{Overview of the two key components of PRENOM.}
    \label{fig_flowchart}
    \vspace{-2pt}
\end{figure*}
Our proposed methodology is composed of two key modules as can be seen in \Cref{fig_flowchart}.
The offline prior trainer (\Cref{sec:offline}) learns a prior only once per category, after which it can be used across all subsequent runs of the online object mapping (\Cref{sec:online}). 
Each mapped object is represented using a compact NeRF model based on the \textit{tiny-cuda-nn} framework\cite{tiny-cuda-nn}, with architecture $\alpha$ and initial parameters $\theta$. 

\subsection{Offline Prior Training}
\label{sec:offline}
We first define a set of categories of interest \(\mathcal{I}\). Then, for each object category \(\mathbf{k} \in \mathcal{I}\), a canonical object frame is defined, and a prior is trained to optimize \(\hat{\alpha}_{\mathbf{k}}\) and \(\hat{\theta}_{\mathbf{k}}\) while generating a mesh \(\overline{\mathcal{M}}_{\mathbf{k}}\) and a prior density grid \(\overline{\mathcal{F}}_{\mathbf{k}}\) in the Normalized Object Coordinate Space (NOCS) \cite{wang2019NOCS}, which is bounded within a unit cube. Priors are trained via meta-learning across several object reconstruction \textit{tasks}.


\smallskip\noindent\textbf{Task Generation }
Each task $\mathcal{T}$ consists of the data required to reconstruct a single object provided its pose and size. This includes a random number of RGB-D frames, instance masks, and camera poses (see \Cref{fig_flowchart} for example), generated using publicly available 3D models primarily from the Shapenet \cite{chang2015shapenet} and Sapiens \cite{Xiang_2020_SAPIEN} datasets. 3D models are divided into training and test set, with $\mathcal{T}^{train}$ used to optimize $\theta$ and $\mathcal{T}^{test}$ used to optimize $\alpha$.

\smallskip\noindent\textbf{Meta-learning}
\begin{figure}[t]
    \centering
    \includegraphics[width=0.85\linewidth]{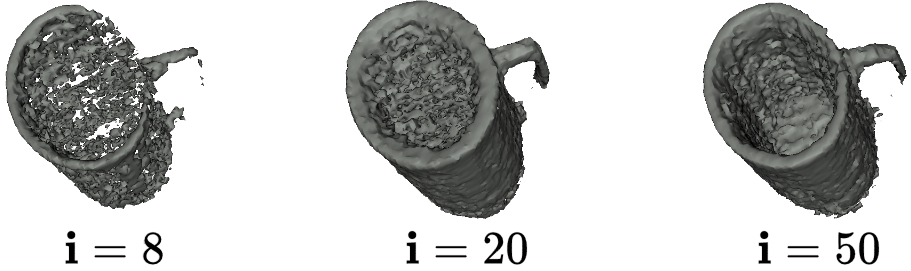}
    \caption{Evolution of the prior normalized mesh \(\overline{\mathcal{M}}_\mathbf{MUG}\) during meta-learning (uniform color for illustration). The process first captures a rough shape, then refines the handle and brim, and finally learns the distinctive cavity of mugs.}
    \label{fig:mug_progress}
\end{figure}
We use the Reptile~\cite{nichol2018reptile} meta-learning algorithm to learn optimized initial parameters $\hat{\theta}$ such that they can be quickly adapted to unseen object instances within the same category during online mapping.
Formally, given a distribution of tasks $\mathcal{T}^{train}_{\mathbf{k}}$ for an object category $\mathbf{k} \in \mathcal{I}$, each task $\mathcal{T}_{\mathbf{k},\mathbf{i}}^{train}$, for the meta-learning step $\textbf{i}$ of $\textbf{N}$, corresponds to learning optimal parameters $\theta_{\mathbf{k}, \mathbf{i}}$ to reconstruct the associated object instance, while maintaining a fixed architecture, by performing gradient descent on the task-specific loss:
\begin{equation}
\theta_{\mathbf{k}, \mathbf{i}}^{\, (t+1)} = \theta_{\mathbf{k}, \mathbf{i}}^{\, (t)} - \eta \nabla_{\theta} \mathcal{L}_{\mathcal{T}}(\theta_{\mathbf{k}, \mathbf{i}}^{\, (t)}),
\label{meta1}
\end{equation}
where $\eta$ is the task learning rate. The loss function $\mathcal{L}_{\mathcal{T}}$ is based on differentiable volume rendering described later in \Cref{sec:online}. After a set number of inner iterations $q$, the updated parameters $\theta_{\mathbf{i}}^{\, (q)}$ are used to update the meta-parameters via a simple interpolation:
\begin{equation}
\theta_{\mathbf{k}, \mathbf{i}+1}^{} = \theta_{\mathbf{k}, \mathbf{i}}^{} + \beta (\theta_{\mathbf{k}, \mathbf{i}}^{\, (q)} - \theta_{\mathbf{k}, \mathbf{i}}^{}),
\label{meta2}
\end{equation}
where $\beta$ is the meta-learning step size. By meta-learning on several reconstruction tasks, the prior learns common geometric features within a category, as seen in \Cref{fig:mug_progress}.

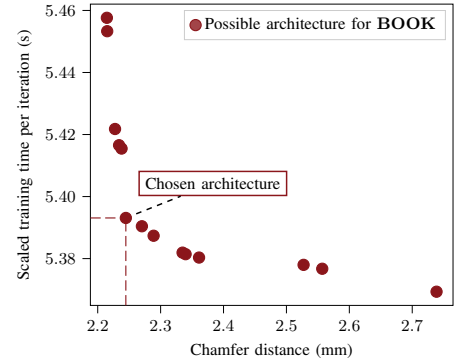
\begin{figure}
    \centering
    \scalebox{0.7}{
\begin{tikzpicture}
\tikzstyle{every node}=[font=\small]

\definecolor{darkgray176}{RGB}{176,176,176}
\definecolor{darkslategray644182}{RGB}{139,22,27}
\definecolor{lightgray204}{RGB}{204,204,204}

\tikzstyle{every pin}=[
    fill=white,
    draw=darkslategray644182,
    font=\small,
] 

\begin{axis}[
legend cell align={left},
legend style={fill opacity=0.8, draw opacity=1, text opacity=1, draw=lightgray204},
tick align=outside,
tick pos=left,
x grid style={darkgray176},
xlabel={Chamfer distance (mm)},
xmin=2.18834736903744, xmax=2.76484554446958,
xtick style={color=black},
y grid style={darkgray176},
ylabel={Scaled training time per iteration (s)},
ymin=5.36491539220251, ymax=5.46204081458315,
ytick style={color=black},
y tick label style={
    /pgf/number format/.cd,
    fixed,
    fixed zerofill,
    precision=2,
    /tikz/.cd
}
]
\addplot [semithick, darkslategray644182, mark=*, mark size=3, mark options={solid}, only marks]
table {%
2.73864108194994 5.36933018412891
2.52726380555978 5.37798553996902
2.55692716030046 5.37673072554614
2.33488155540302 5.38192447475055
2.36109205993195 5.38034001972439
2.33968876909366 5.38142738838024
2.27023169698047 5.39042815001376
2.24470940918646 5.39310578920175
2.28885012309049 5.38737074986879
2.21508124744497 5.45330983794272
2.21455183155708 5.45762602265676
2.23395264196921 5.41655236974493
2.22752678638523 5.42178572027533
2.23789613448127 5.4154758500881
};
\node [coordinate, pin={[pin edge={black, dashed, thick, -}, distance=25pt, thick, align=left]60:{Chosen architecture}}] 
      at (axis cs:2.24470940918646, 5.39310578920175) {};

\addlegendentry{Possible architecture for $\mathbf{BOOK}$}
\addplot [semithick, darkslategray644182, dash pattern=on 5.55pt off 2.4pt, forget plot]
table {%
2.18834736903744 5.39310578920175
2.24470940918646 5.39310578920175
};
\addplot [semithick, darkslategray644182, dash pattern=on 5.55pt off 2.4pt, forget plot]
table {%
2.24470940918646 5.36491539220251
2.24470940918646 5.39310578920175
};
\end{axis}

\end{tikzpicture}}
    \caption{Set of optimal architectures for category $\mathbf{BOOK}$. The final architecture (dashed lines) is chosen as the knee point of the convex decreasing curve.}
    \label{fig:knee}
\end{figure}

\smallskip\noindent\textbf{Architecture Optimization }
To choose $\hat{\alpha}$, we perform a multi-objective optimization, using a mixed-variable genetic algorithm~\cite{pymoo} with the objectives of minimizing the training time and the object reconstruction error (measured using Chamfer distance).
At each optimization iteration, meta-learning is used to obtain the optimized parameters $\theta_{\mathbf{k}, \mathbf{N}}$, as described in \Cref{meta1,,meta2}. These learned parameters are then evaluated on $\mathcal{T}^{test}_{\mathbf{k}}$ to guide the genetic algorithm to generate an improved architecture for the next iteration.
Starting with an initial population of architectures with size $\mathbf{P}$, this iterative process is repeated for $\mathbf{M}$ generations, producing a set of Pareto-optimal solutions. 
From these solutions, the final architecture \(\hat{\alpha}_{\mathbf{k}}\) is chosen as the knee point of the convex decreasing curve that best fits the solution set (see \Cref{fig:knee}).  
Consequently, $\hat{\theta}_{\mathbf{k}}$ are the corresponding meta-learned parameters for the selected architecture.


\smallskip\noindent\textbf{Normalized mesh generation }
$\overline{\mathcal{M}}_{\mathbf{k}}$ is generated using marching cubes on the density output of the NeRF with the optimized initial parameters $\hat{\theta}_{\mathbf{k}}$. The density output, computed over a cubic grid of resolution $R$, is stored as $\overline{\mathcal{F}}_\mathbf{k}$. During online object mapping, the mesh $\overline{\mathcal{M}}$ and density grid $\overline{\mathcal{F}}$ are used to canonicalize object poses.

\subsection{Online Object Mapping}
\label{sec:online}
Our mapping framework builds on top of RO-MAP \cite{han2023ro}, which extends ORB-SLAM2 \cite{mur2017orb2}, maintaining real-time capability. 
Camera poses are directly sourced from ORB-SLAM2’s tracking module.

\smallskip\noindent\textbf{Object Association } Our association strategy is adapted from \cite{han2023sq, wu2023object}. For consecutive frames, objects are considered the same if their bounding box Intersection over Union (IoU) exceeds a predefined threshold, while for non-consecutive frames, projected IoU (p-IoU) is used. 
If these criteria are met, we refine the association by using object point clouds from depth frames and instance masks. Each incoming object cloud is downsampled and clustered, with small clusters treated as outliers. The remaining clusters undergo statistical testing. Since object clouds do not follow a Gaussian distribution, we apply the Wilcoxon Rank-Sum Test \cite{wilcoxon1992individual} to assess distribution similarity. Meanwhile, object cloud centroids are assumed to follow a Gaussian distribution, verified using a one-sample t-test. If both tests pass, the incoming object cloud is merged with the existing one.

\smallskip\noindent\textbf{Coarse Object State Estimation } The 9-DoF state comprises a 6-DoF pose and a 3-DoF size. Consistent with related works \cite{han2023ro, wu2023object}, we assume that objects are parallel to the ground, reducing the pose estimation problem to position and yaw estimation. The optimal yaw is determined by aligning line features extracted from images with the edges of the bounding box. Once yaw is estimated, the object size is computed from the lengths of its spatial extents, while its position is defined as the center of these extents.

\smallskip\noindent\textbf{Object State Canonicalization }
\begin{figure}[t]
    \centering
    \includegraphics[width=0.75\linewidth]{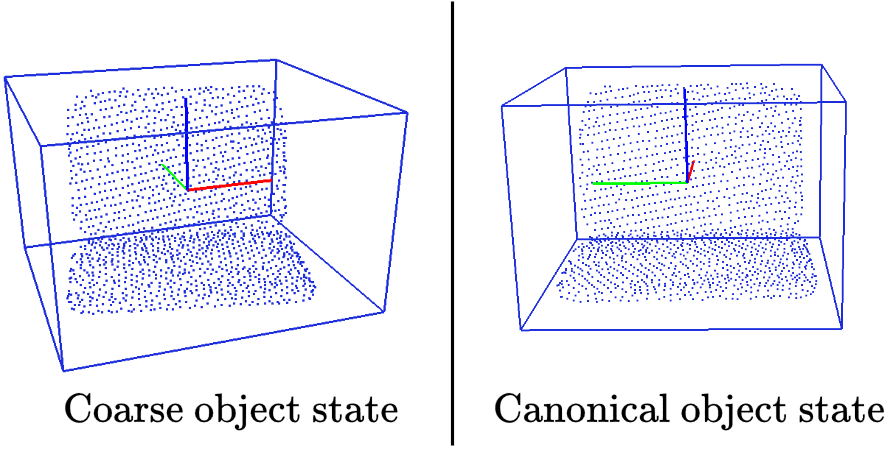}
    \caption{Before and after pose canonicalization. Our approach aligns the object frame with the canonical frame established during offline prior training ($x$-axis, in red, points into the laptop screen), which also refines the fit of the bounding box.}
    \label{fig:laptop_canon}
\end{figure}
An object is considered ready for reconstruction after it has been observed from a predefined sufficiently wide range of viewpoints. If the object has a prior (semantic class \(\mathbf{k} \in \mathcal{I}\)), its pose is transformed into the canonical frame relative to which the prior was learned. For this, we sample yaw angles between \(0\) and \(2\pi\), and for each sample, apply the corresponding yaw rotation to the object point cloud \(P\) to obtain \(P_\gamma\). The object's size \(\mathcal{S}_\gamma\) is estimated as the spatial extents of \(P_\gamma\) and is used to normalize the point cloud, producing \(\overline{P}_\gamma = P_\gamma / \mathcal{S}_\gamma + 1/2\) leading to \(\overline{P}_\gamma \in [0, 1]\).
The optimal yaw angle $\hat{\gamma}$ is then selected by maximizing the accumulated density of the normalized points in the prior density grid:
\begin{equation}
    \hat{\gamma} = \argmax_\gamma \sum_{j=1}^{\left| P \right|} \,\overline{\mathcal{F}}_{\mathbf{k}}(\overline{P}_{\gamma,j})
\label{eq:canonical}
\end{equation}
where $\overline{\mathcal{F}}_{\mathbf{k}}(x)$ represents the density of a normalized point $x$, computed via trilinear interpolation over the fixed-resolution $\overline{\mathcal{F}}_{\mathbf{k}}$. Here, $\overline{P}_{\gamma,j}$ denotes the $j^{\,\text{th}}$ point of the normalized object cloud $\overline{P}_{\gamma}$, which contains $\left| P \right|$ points. Note that this optimization is unnecessary for objects with rotational symmetry.
Moreover, as a final step, we transform and scale $\overline{\mathcal{M}}_{\mathbf{k}}$ using $\hat{\gamma}$ and $\mathcal{S}{\hat{\gamma}}$ to obtain a reference mesh, which is then aligned using ICP.

This process not only ensures that the prior is loaded correctly but also refines the object state, resulting in a compact and well-aligned bounding box, as shown in \Cref{fig:laptop_canon}.





\smallskip\noindent\textbf{Object-level NeRFs }
Once the object state is estimated, it is implicitly reconstructed using a lightweight NeRF model, which incorporates a multi-resolution hash encoding~\cite{mueller2022instant} alongside a shallow MLP.
Supervision is achieved through differentiable volume rendering, which samples and accumulates color and opacity along camera rays to reconstruct the object. For a known object $\mathbf{k} \in \mathcal{I}$, the optimized initial parameters $\hat{\theta}_{\mathbf{k}}$ are loaded into a NeRF model with architecture $\hat{\alpha}_{\mathbf{k}}$. Otherwise, a default architecture (adapted from \cite{han2023ro}) and random parameter initialization are used.
Refer to \Cref{fig_flowchart} for color-coded flow. 

\smallskip\noindent\textbf{Training Data } RGB-D images, instance masks, and camera poses at every keyframe are used to train the NeRFs. 

\smallskip\noindent\textbf{Probabilistic Ray Sampling } We employ a ray sampling strategy based on a Cumulative Distribution Function (CDF) derived from $\overline{\mathcal{F}}$, which steers sampling towards regions with high occupancy probability. 
\begin{figure}[t]
    \centering
    \includegraphics[width=0.75\linewidth]{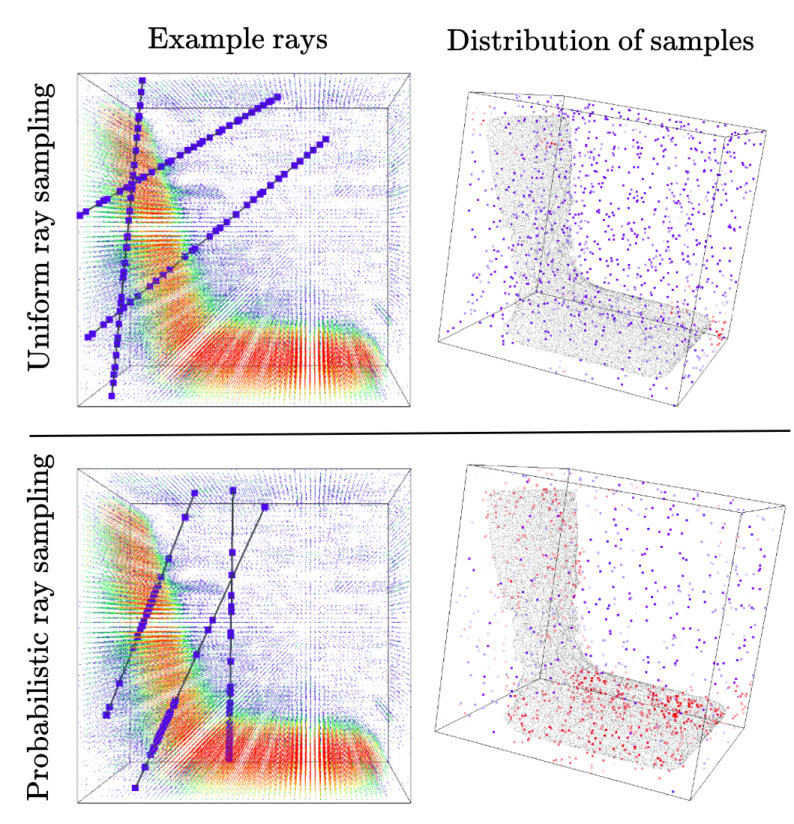}
    \caption{Comparison of ray sampling strategies: Uniform vs. Probabilistic (Ours). \textit{Left}: Three random rays with sampled points overlaid on $\overline{\mathcal{F}}_{\mathbf{LAPTOP}}$. \textit{Right}: Distribution of sampled points with red denoting high density. Probabilistic sampling focuses more points where the mesh is expected.}
    \label{fig:ray_sampling}
\end{figure}
All camera rays are first transformed into the object frame using the estimated object pose, and \(N_c\) samples are drawn at equal distance intervals along the ray, bounded within the object's bounding box. These serve as the discrete intervals to estimate the occupancy probability given by, 
\begin{equation}
\rho_\text{occ}(\mathbf{x}_i) = 1 - \exp(-\sigma_i\delta_i)
\end{equation}
where $\delta_i=\mathbf{d}_{i+1}-\mathbf{d}_i$, is the distance between adjacent sampled points $\mathbf{x}_{i+1}$ and $\mathbf{x}_i$. For ray sampling, we set $\sigma_i = \overline{\mathcal{F}}(\mathbf{x}_i)$ to both provide an initial estimate and avoid costly inference. The termination probability then becomes,
\begin{equation}
    \rho_{\text{term}}(\mathbf{x}_i) = \rho_{\text{occ}}(\mathbf{x}_i) \prod_{j=1}^{i-1} (1 - \rho_{\text{occ}}(\mathbf{x}_j))
\label{term_eq}
\end{equation}
using which a discrete CDF is constructed as follows:
\begin{equation} \text{CDF}(i) = \frac{\sum_{j=1}^{i} \rho_{\text{term}}(\mathbf{x}_j)}{\sum_{j=1}^{N_c} \rho_{\text{term}}(\mathbf{x}_j)}\end{equation}
This monotonically increasing function represents the probability that the ray terminates at or before each sampled point. Finally, we use inverse transform sampling on the discrete CDF distribution to generate the $N_r$ sampled points
$\mathbf{p}_k$ for ray $\mathbf{r}_k$. Doing so 
ensures that more samples are concentrated in regions where the mesh is expected, as can be seen in \Cref{fig:ray_sampling}. For rays that \textit{escape} ($\sum_{i=1}^{N} \rho_{\text{term}}(\mathbf{x}_i) < \epsilon$, where $\epsilon$ is a small number), we revert to a uniform sampling strategy, maintaining coverage in low-density regions. Moreover, as training progresses, $\overline{\mathcal{F}}$ is updated every $m$ iterations through the mesh rendering process. This ensures that sampling progressively converges on the actual object, even if the initial object pose detection was imprecise.


\smallskip\noindent\textbf{Volume Rendering } An overview of the differentiable volume rendering \cite{kong2023vmap, abou2022implicit, han2023ro} is provided for completeness.
For each object-centric ray $\mathbf{r}$, the positions of the sampled points $\mathbf{p}$ are normalized and encoded, then provided as input to an MLP to obtain the density $\sigma$ and the RGB color $\mathbf{c}$ for all points. The predicted color and depth of the ray are then accumulated using \Cref{term_eq} as follows:
\begin{equation}
    \hat{\mathbf{C}}(\mathbf{r}) = \sum_{j=1}^{N_r} \rho_{\text{term}}(\mathbf{p}_j) \mathbf{c}_j, \;\; \hat{\mathbf{D}}(\mathbf{r})= \sum_{j=1}^{N_r} \rho_{\text{term}}(\mathbf{p}_j) \mathbf{d}_j
\end{equation}  
To compute the loss in the forward pass, rays $\mathcal{R}$ are categorized into those hitting the object and those that are part of the background as $\mathcal{R}_o$ and $\mathcal{R}_b$ respectively. Then the photometric loss $\mathcal{L}_{\mathbf{c}}$, depth loss $\mathcal{L}_{\mathbf{d}}$, and density loss $\mathcal{L}_{\sigma}$ are defined as:  
\begin{equation}
\begin{gathered}
    \mathcal{L}_{\mathbf{c}} = \sum_{k \in \mathcal{R}_o} \norm{\hat{\mathbf{C}}(\mathbf{r}_k)-\mathbf{C}(\mathbf{r}_k)}_2 + \sum_{k \in \mathcal{R}_b} \norm{\hat{\mathbf{C}}(\mathbf{r}_k)-\mathbf{C}_{\text{rand}}}_2 \\
    \mathcal{L}_{\mathbf{d}} = \sum_{k \in \mathcal{R}_o} \abs{\hat{\mathbf{D}}(\mathbf{r}_k)-\mathbf{D}(\mathbf{r}_k)}, \;\;  \mathcal{L}_{\sigma} = \sum_{k \in \mathcal{R}_b} \sum_{j=1}^{N_r} \abs{\sigma_{k,j}}
\end{gathered}  
\end{equation}
where $\sigma_{k,j}$ is the predicted density for the $j^{\,\text{th}}$ sampled point of the $k^{\,\text{th}}$ ray. Here, $\mathbf{C}_{\text{rand}}$ is random color supervision that, together with $L_{\sigma}$, guide the network to learn zero densities for points sampled in the background of the object, as done in \cite{han2023ro}. The final loss $L$ is a weighted sum,  
\begin{equation}
    \mathcal{L} = \mathcal{L}_{\mathbf{c}} + \lambda_{\mathbf{d}} \mathcal{L}_{\mathbf{d}} + \lambda_{\sigma} \mathcal{L}_{\sigma}
\end{equation}  
which is back-propogated to optimize the NeRF parameters using the Adam optimizer.

\section{Evaluation}
\label{sec_evaluation}
\subsection{Experimental Setup and Implementation Details}
To train piors, we run the genetic algorithm, to optimize the architectures $\{\alpha_\mathbf{k}\,|\,\mathbf{k}\in\mathcal{I}\}$, with initial population size $\mathbf{P}\mathbin{=}25$ and for $\mathbf{M}\mathbin{=}100$ generations. Each $\alpha$ consists of all tunable hyperparameters of the NeRF representation (refer to \cite{mueller2022instant} for a full list), the meta-learning ($\mathbf{N}$, $q$, $\eta$, and $\beta$), and the volume rendering loss ($\lambda_\mathbf{d}$ and $\lambda_\sigma$). 
For object reconstruction during online mapping, 4096 rays are generated at each iteration with $N_c\mathbin{=}N_r\mathbin{=}32$ samples per ray. Moreover, we set $R\mathbin{=}64$, $\epsilon\mathbin{=}10^{-4}$ and $m\mathbin{=}50$. 
The probabilistic ray sampling is implemented using CUDA kernels ensuring minimal computational overhead. 
Moreover, a new thread is created for the training of each object-level NeRF.

We run the experiments on a system equipped with an Intel Core i9-12900H and a low-power, low-memory (4GB) NVIDIA T600 Laptop GPU. 
With this setup, only a limited number of training iterations per object is possible to maintain sequential real-time processing of multi-object scenarios.


\subsection{Comparison with prior-free methods}

\begin{figure}[t]
    \vspace{5pt}
    \centering
    \begin{minipage}{0.41\textwidth}
        \centering
        \includegraphics[width=0.40\linewidth]{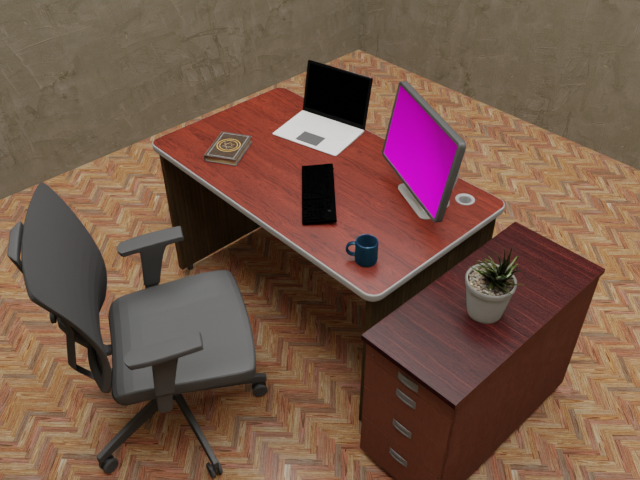}
        \centering
        \includegraphics[width=0.40\linewidth]{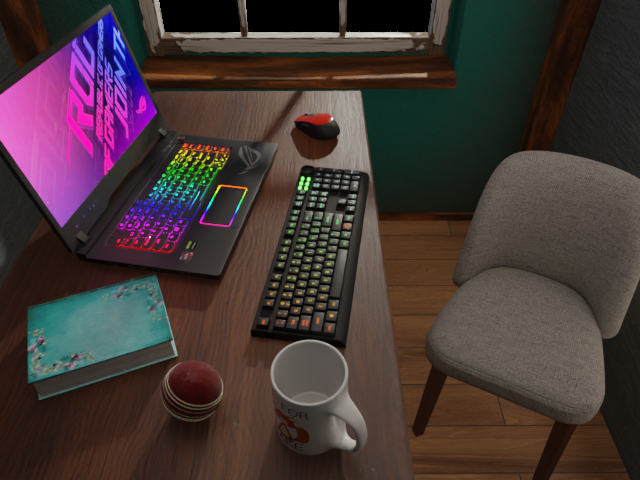}

        \vspace{0.015\linewidth}
                
        \centering
        \includegraphics[width=0.40\linewidth]{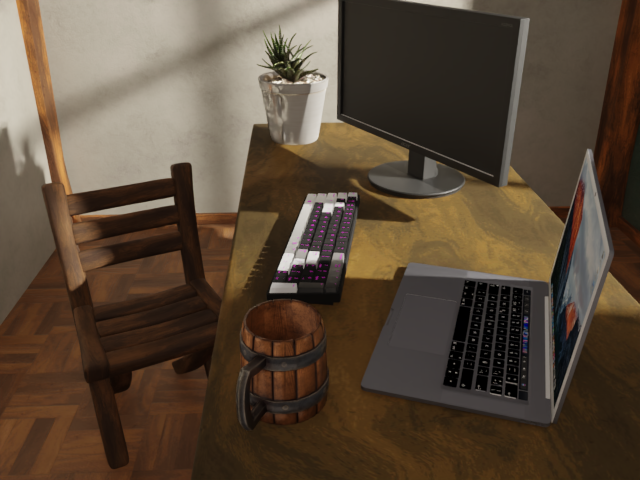}
        \centering
        \includegraphics[width=0.40\linewidth]{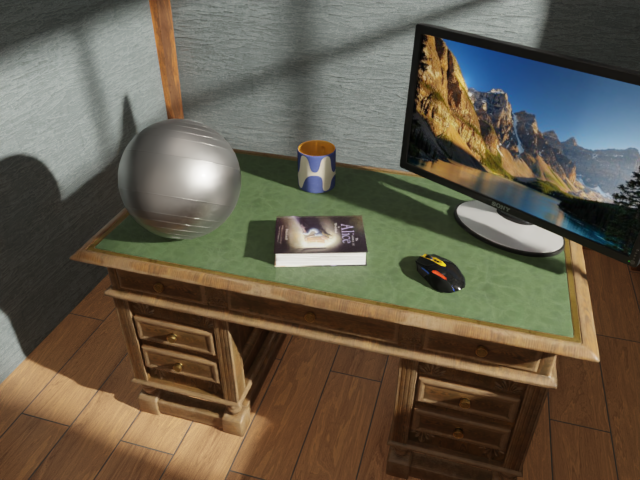}
    \end{minipage}
    \caption{Example scenes from the generated synthetic dataset.}
    \label{fig:example_simulated}
\end{figure}

\begin{table*}[t]
    \centering
    \vspace{5pt}
    \begin{tabular}{l|
    >{\centering\arraybackslash}p{0.5cm}
    >{\centering\arraybackslash}p{0.5cm}
    >{\centering\arraybackslash}p{0.5cm}
    >{\centering\arraybackslash}p{0.5cm}
    >{\centering\arraybackslash}p{0.5cm}
    >{\centering\arraybackslash}p{0.5cm}
    >{\centering\arraybackslash}p{0.5cm}
    >{\centering\arraybackslash}p{0.5cm}
    >{\centering\arraybackslash}p{0.5cm}
    >{\centering\arraybackslash}p{0.5cm}
    >{\centering\arraybackslash}p{0.5cm}
    >{\centering\arraybackslash}p{0.5cm}|
    >{\centering\arraybackslash}p{0.5cm}
    >{\centering\arraybackslash}p{0.5cm}|
    >{\centering\arraybackslash}m{0.5cm}}
        \toprule
        \multirow{2}{*}{Method} & \multicolumn{2}{c}{S0 (4 objects)} & \multicolumn{2}{c}{S1 (5 objects)} & \multicolumn{2}{c}{S2 (5 objects)} & \multicolumn{2}{c}{S3 (5 objects)} & \multicolumn{2}{c}{S4 (6 objects)} & \multicolumn{2}{c}{S5 (7 objects)} & \multicolumn{2}{c}{\textbf{Average/obj.}} & \multirow{1}{0.5cm}{\parbox{1\linewidth}{\vspace{0.2cm} Time (s)↓}}\\
        \cmidrule(lr){2-13} \cmidrule{14-15} 
        & CD↓ & CR↑ & CD↓ & CR↑ & CD↓ & CR↑ & CD↓ & CR↑ & CD↓ & CR↑ & CD↓ & CR↑ & CD↓ & CR↑ \\
        \midrule
        vMap & 0.538 & 0.396 & 0.718 & 0.283 & 0.985 & 0.263 & \underline{0.615} & 0.481 & \underline{1.404} & 0.233 & \underline{1.122} & 0.337 & \underline{0.870} & 0.334 & 19.9 \\
        \hspace{-1pt}\cite{abou2022implicit}* & 0.724 & 0.518 & 0.920 & 0.405 & 1.434 & 0.308 & 1.062 & 0.387 & 1.944 & 0.197 & 1.933 & 0.260 & 1.282 & 0.358 & 24.0 \\
        RO-MAP* & \underline{0.434} & \textbf{0.705} & \underline{0.657} & \underline{0.488} & \underline{0.917} & \textbf{0.378} & 0.850 & \underline{0.656} & 1.523 & \underline{0.255} & 1.167 & \textbf{0.482} & 0.886 & \underline{0.504} &  19.7\\
        \midrule
        \textit{Ours} & \textbf{0.375} & \underline{0.669} & \textbf{0.541} & \textbf{0.527} & \textbf{0.811} & \underline{0.351} & \textbf{0.608} & \textbf{0.724} & \textbf{1.127} & \textbf{0.279} & \textbf{0.785} & \underline{0.477} & \textbf{0.688} & \textbf{0.512} & 17.9\\
        \bottomrule
    \end{tabular}
    \caption{Evaluation results for six sequences with increasing number of objects. CD represents the Chamfer Distance (in cm) and CR represents the Completion Ratio ($<\,$0.4\,cm). All experiments were repeated thrice and the mean reported.}
    \label{tab:effective}
\end{table*}

We assess effectiveness by comparing against prior-free methods that also use NeRFs as the object representation.

\smallskip\noindent\textbf{Baselines } We compare against three approaches: vMap \cite{kong2023vmap}, \cite{abou2022implicit}, and RO-MAP \cite{han2023ro}. For vMAP, we provide its open-source implementation with poses from ORB-SLAM2 and disable the scene background reconstruction. For RO-MAP, we extend their monocular framework by incorporating depth data, which fixes scale ambiguity and enables dense depth supervision. We refer to this modified version as RO-MAP*. For \cite{abou2022implicit}, since it is not open-source, we adapt it from RO-MAP* by using the architecture and hyperparameters mentioned in their paper, naming it \cite{abou2022implicit}*. For all comparisons, we adjust the training iterations to maintain sequential real-time multi-object mapping. However, since training is performed in batches, we allow the last batch in the queue to complete. Moreover, if a framework produces an empty object mesh, we increase the training iterations accordingly. Therefore for fairness, we also report the average processing time per sequence. Notably, all frameworks operate in a multi-object setting, training objects in parallel.

\smallskip\noindent\textbf{Dataset } We evaluate on a synthetic dataset with available ground-truth meshes by extending the Cube Diorama dataset \cite{abou2022implicit}. Originally comprising a single scene with four objects, we expand it to six scenes featuring 32 unique objects across nine diverse categories: \textit{display, laptop, chair, mug, ball, keyboard, mouse, book}, and \textit{plant} (see \Cref{fig:example_simulated} for examples). Notably, none of the object instances in this dataset were encountered by PRENOM during the offline prior training phase. Moreover, instance masks are provided offline to ensure consistent input data across all frameworks.

\smallskip\noindent\textbf{Metrics } We use Chamfer Distance (CD) and Completion Ratio (CR) to quantitatively evaluate reconstruction quality. We do not evaluate object state recovery since prior-free methods cannot localize objects in a canonical frame.

\begin{figure}[t]
    \centering
    \includegraphics[width=\linewidth]{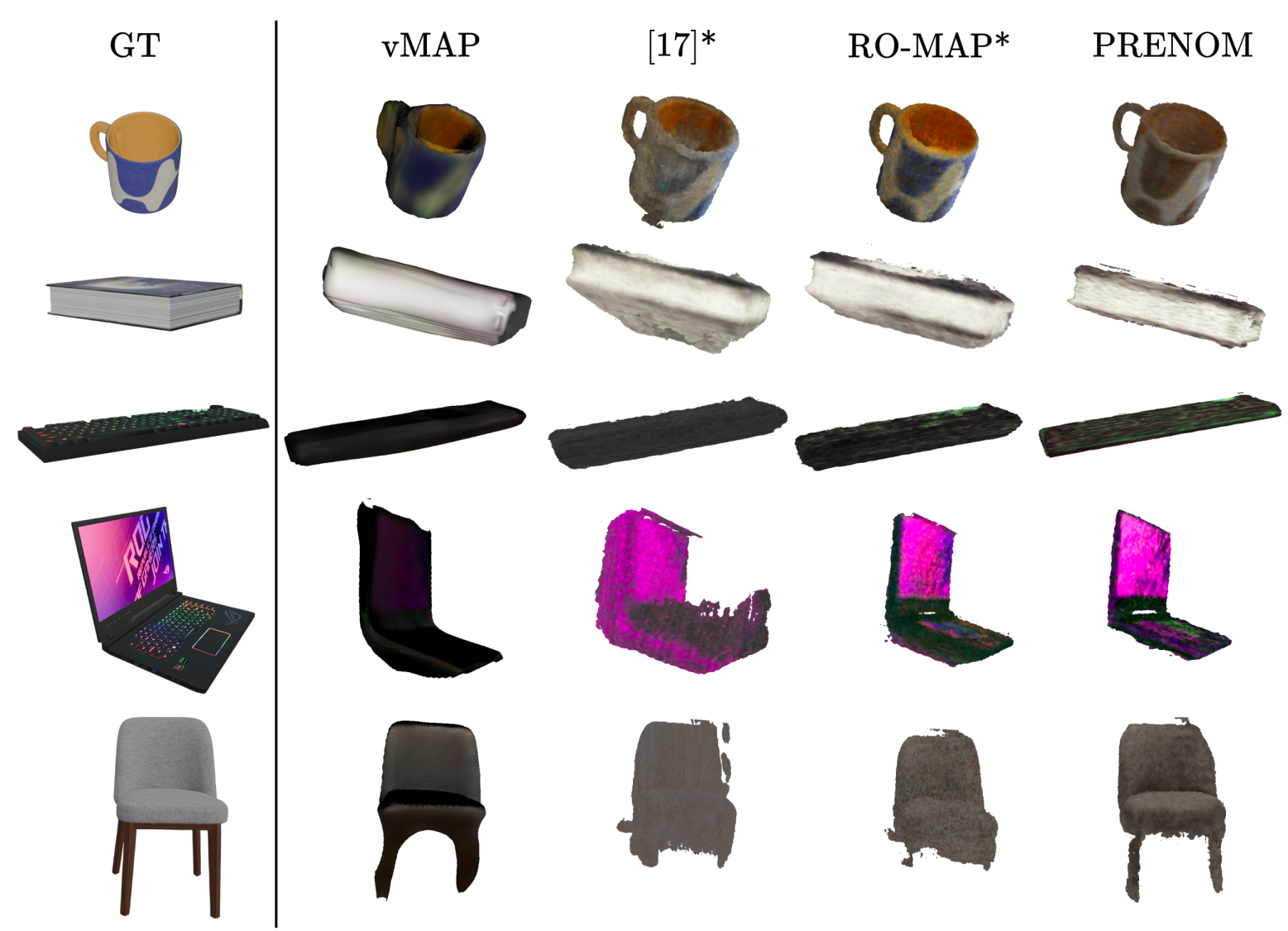}
    \caption{PRENOM generates more refined and detailed geometry, such as the brim and cavity of the mug, pages of the book, flat surfaces of the keyboard and laptop, and thin legs of the chair. vMap produces over-smoothed geometry, while \cite{abou2022implicit}* and RO-MAP* exhibit excess geometry.}
    \label{fig:qual1}
\end{figure}

\smallskip\noindent\textbf{Results and discussion } Quantitative analysis in \Cref{tab:effective} shows that PRENOM consistently achieves a significantly lower CD across all scenes, outperforming the next-best method by 21\% on average, while also achieving a slight improvement in average CR. Qualitative results on example objects in \Cref{fig:qual1} further illustrate PRENOM’s ability to reconstruct sharper geometry, such as the brim and cavity of a mug, the flat surfaces of a book, keyboard, and laptop, and the thin legs of a chair. In contrast, excess geometry generated by RO-MAP* contributes to its improved completion ratio in some scenes, as evidenced by its significantly lower CD. It is also worth noting that PRENOM has been optimized to prioritize accurate shape reconstruction over photometric quality, which can result in minor losses in visual fidelity (e.g., the blue in the mug). This is a reasonable trade-off for object mapping scenarios, where novel-view rendering is typically unimportant. However, if needed, our prior generation pipeline can accommodate additional optimization objectives to improve photometric consistency. 

\subsection{Comparison with category-level shape prior approaches}
Most open-source approaches rely on DeepSDF as a shape prior which is computationally expensive, particularly in multi-object scenarios on a low-end GPU. Therefore, we run these evaluations in an offline, single-object setting.

\smallskip\noindent\textbf{Baselines } We compare against three approaches: NodeSLAM~\cite{sucar2020nodeslam}, DSP-SLAM~\cite{wang2021dsp}, and UncertainShapePose (USP)~\cite{liao2024uncertainshapepose}. To ensure a fair object-mapping-only comparison, we adapt all approaches from the open-source implementation of USP. NodeSLAM does not originally use DeepSDF as a prior but since is not open-source, we use the variant of USP with the uncertainty component and 3D loss disabled, incorporating the sampling-based rendering from NodeSLAM while retaining the energy score from USP, denoted as NodeSLAM*. For DSP-SLAM, we use the variant of USP with the uncertainty component disabled and deterministic 3D and 2D loss functions, denoted as DSP-SLAM*. For the original USP, we retain the uncertainty component as it is integrated with the optimization process, despite the increased computational cost. Additionally, we also evaluate against a prior-free variant of PRENOM. This variant, while using the same initial object state and architecture as PRENOM, reconstructs an object-level NeRF from scratch without using meta-learned parameters.

For efficient execution of all frameworks, we analyze the loss curves of each method to determine the average number of iterations required for convergence. This iteration count is then used as the stopping criterion to ensure near-optimal results while maintaining computational efficiency. Therefore for reference, we also record the average runtime per object provided for each framework, though it does not necessarily represent the minimum time required.


\smallskip\noindent\textbf{Dataset } We evaluate on the real-world ScanNet~\cite{dai2017scannet} dataset, focusing on the \textit{chair} category. The dataset provides noisy depth and instance masks, while reference meshes are sourced from Scan2CAD~\cite{2019scan2cad} annotations. Ground-truth camera poses are used for all methods. We select 10 scenes containing 43 chair instances to assess both reconstruction quality and object state recovery. For all methods, we use the same 25 views, with identical segmentation masks and initial estimates of the object state. 

\smallskip\noindent\textbf{Metrics } To evaluate reconstruction quality, we use CD and CR as before. Moreover, we also incorporate Earth Mover's Distance (EMD) to provide a robust assessment of global shape, since NeRF-based approaches tend to produce sharper local features than DeepSDF-based approaches, and CD alone can be biased toward these details. 

For object state estimation, we compute translation, rotation, and size errors. Morover, similar to PRENOM, the baseline methods also assume that chairs lie parallel to the ground to compute the initial pose. However, since these methods update the pose during reconstruction, we align their final poses' vertical rotation axes with the ground truth before evaluation to ensure fairness.
In contrast, PRENOM only estimates the pose using $\overline{\mathcal{F}}_\mathbf{CHAIR}$ and $\overline{\mathcal{M}}_\mathbf{CHAIR}$ (\Cref{eq:canonical} for reference) once before reconstructing, so it's vertical rotation axis is already aligned.

\smallskip\noindent\textbf{Results and discussion } PRENOM achieves an improvement in reconstruction quality of approximately 13\% averaged across all metrics compared to the second-best approach, as shown in \Cref{tab:quant_rec_prior}. Notably, it outperforms other methods in reconstruction quality while only being trained for 5$\times$ less time and with a 30\% smaller prior size. Additionally, comparisons with our prior-free version re-emphasize that our prior-based approach not only enables canonical object state estimation but also significantly enhance reconstruction quality.

\begin{table}[t]
    \vspace{5pt}
    \centering
    \begin{tabular}{p{1.85cm}|
    >{\centering\arraybackslash}p{1.0cm}
    >{\centering\arraybackslash}p{1.0cm}
    >{\centering\arraybackslash}p{1.0cm}|
    >{\centering\arraybackslash}p{0.5cm}
    >{\centering\arraybackslash}p{0.5cm}}
        \toprule
        \multirow{2}{1.5cm}{Method} & CD↓ & EMD↓ & CR↑ & Size↓ & Time↓  \\
        &(cm) &(cm) & ($<$1\,cm) & (MB) & (s)\\
        \midrule
        NodeSLAM* & 6.927  & 8.995& 0.157 & 25.6 & 45.3 \\
        DSP-SLAM* & \underline{6.136}  & 9.050& 0.179 & 25.6 & 46.3 \\
        USP \cite{liao2024uncertainshapepose} & 6.301 & \underline{8.859} & 0.189 & 25.6 & 109.9 \\
        \midrule
        \textit{Ours} (w/o prior) & 6.273  & 10.024 & \underline{0.254} & - & 9.4 \\
        \textit{Ours} & \textbf{5.189}  & \textbf{8.017}& \textbf{0.294} & 17.9 & 9.1 \\
        \bottomrule
    \end{tabular}
    \caption{Quantitative comparison of object reconstruction performance of shape prior approaches. Size represents the disk space required to store the prior, including $\overline{\mathcal{F}}_\mathbf{CHAIR}$ and $\overline{\mathcal{M}}_\mathbf{CHAIR}$ for PRENOM. Time indicates the reconstruction time provided per object (not minimum).}
    \label{tab:quant_rec_prior}
\end{table}

\begin{figure}[t]
    \centering
    \includegraphics[width=0.93\linewidth]{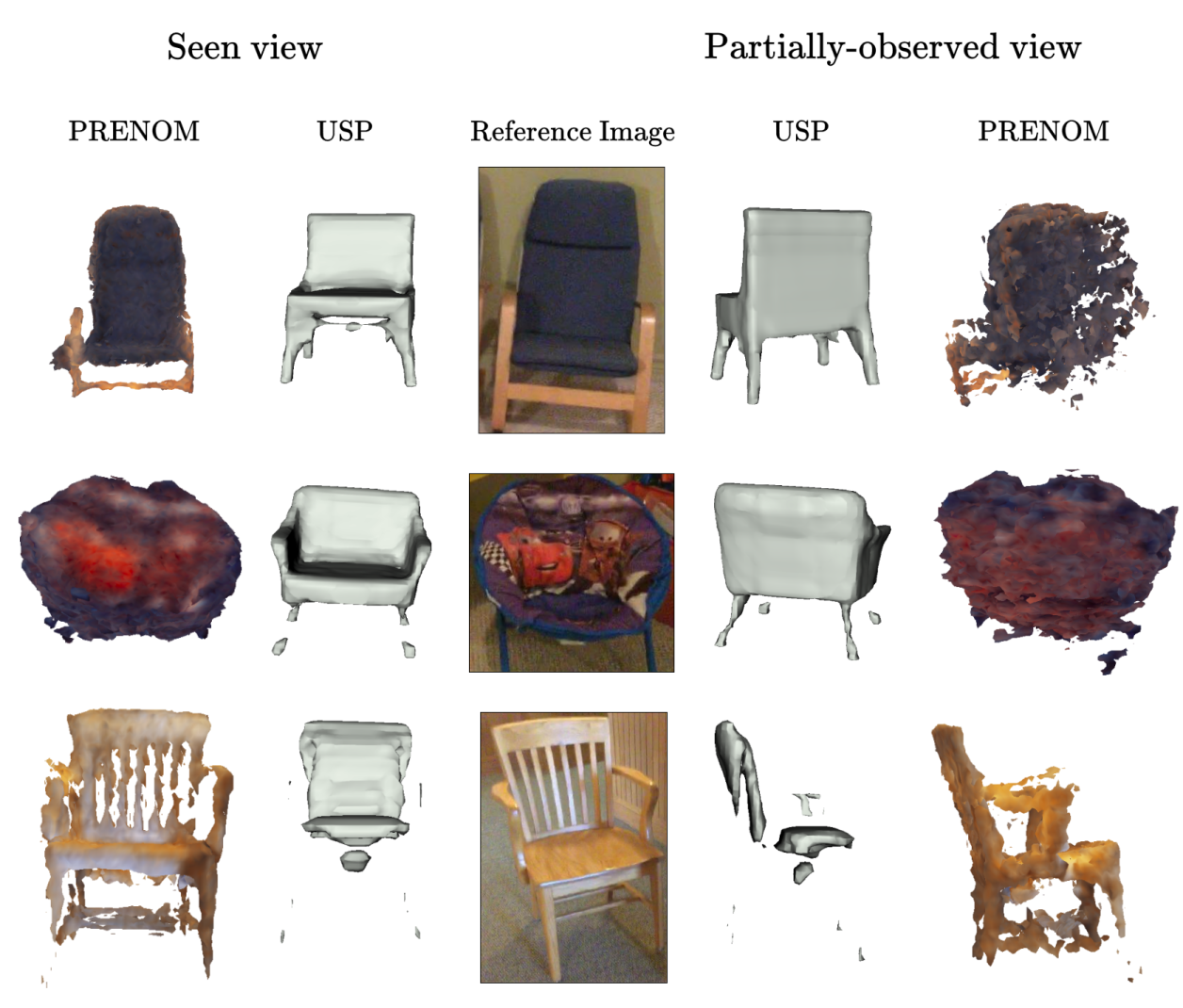}
    \caption{PRENOM generates sharper geometry than the longest trained DeepSDF-based method (USP) for in-view regions but struggles with out-of-view features as scene information eventually dominates the weak prior. Both methods struggle with legs because of very noisy instance segmentation. Reference image is subjectively the clearest one of 25 input views.}
    \label{fig:qual2}
\end{figure}

Furthermore, example qualitative results in \Cref{fig:qual2} demonstrate PRENOM's ability to reconstruct sharp geometrical features within the observed regions. However, we do not impose geometric constraints with the initial prior, and during training, scene information gradually dominates, leading to noisier reconstructions in partially-observed/unobserved regions and regions with occlusions. In contrast, DeepSDF-based methods struggle to reconstruct complex and thin geometries typical of chairs but can predict the unseen regions in a way that remains structurally similar to a chair, even if it is not consistent with the scene. 

\begin{table}[t]
    \vspace{5pt}
    \centering
    \begin{tabular}{l|ccc}
        \toprule
        Method & Trans. (cm)↓ & Rot. (\textdegree)↓ & Size (\%)↓  \\
        \midrule
        NodeSLAM* & \textbf{14.38}  & 52.58 & 36.27  \\
        DSP-SLAM* & \underline{14.74}  & 52.48 & \underline{24.48} \\
        USP \cite{liao2024uncertainshapepose} & 14.95 & 50.88 & \textbf{23.41} \\
        \midrule
        \textit{Ours} & 18.60  & 19.46 & 31.60 \\
        \bottomrule
    \end{tabular}
    \caption{Quantitative comparison with other shape prior approaches for object pose and size estimation.}
    \label{tab:quant_pose_prior}
\end{table}


Moreover, despite not jointly optimizing object state with reconstruction, our method achieves comparable object state recovery results as can be seen in \Cref{tab:quant_pose_prior}. We do not consider the improvement in rotation accuracy to be significant, as PRENOM does not adjust pose during reconstruction and thus benefits from the initial assumption that objects are parallel to the ground. While this initial assumption is used by all methods, the baseline approaches are more susceptible to noise as they alter the pose during the optimization process, which can negatively impact the rotation accuracy. 

\subsection{Ablation Study}

\begin{table}[t]
    \centering
    \begin{tabular}{p{0.7cm}|
    >{\centering\arraybackslash}p{0.25cm}
    >{\centering\arraybackslash}p{0.25cm}
    >{\centering\arraybackslash}p{0.25cm}
    ccc}
        \toprule
        \multirow{2}{0.2cm}{\centering Mode} & \multirow{2}{0.2cm}{\centering AO} & \multirow{2}{0.2cm}{\centering ML} & \multirow{2}{0.2cm}{\centering PS} & CD↓ & CR↑ & Size↓ \\
        &&&& (cm) & ($<$0.4\,cm) & (MB)\\
        \midrule
        \multirow{4}{0.7cm}{\centering{\rotatebox[origin=c]{90}{RGB}}}
        & \XSolidBrush & \XSolidBrush & \XSolidBrush & 0.495 & 0.545 & 17.2 \\
        & \Checkmark & \XSolidBrush & \XSolidBrush & 0.422 (\cellcolor{yellow}{↓ 15\%}) & 0.600 (\cellcolor{yellow}{↑ 10\%}) & 5.5 \\
        & \Checkmark & \Checkmark & \XSolidBrush & 0.369 (\cellcolor{greenl}{↓ 25\%}) & 0.660 (\cellcolor{greenl}{↑ 21\%}) & 5.5 \\
        & \Checkmark & \Checkmark & \Checkmark & \textbf{0.330 (\cellcolor{greend}{↓ 33\%})} & \textbf{0.732 (\cellcolor{greend}{↑ 34\%})} & 5.5\\
        \midrule
        \multirow{4}{0.7cm}{\centering{\rotatebox[origin=c]{90}{RGB-D}}}
        & \XSolidBrush & \XSolidBrush & \XSolidBrush & 0.455 & 0.569 & 17.2 \\
        & \Checkmark & \XSolidBrush & \XSolidBrush & 0.368 (\cellcolor{yellow}{↓ 19\%}) & 0.630 (\cellcolor{yellow}{↑ 11\%}) & 5.5 \\
        & \Checkmark & \Checkmark & \XSolidBrush & 0.349 (\cellcolor{greenl}{↓ 23\%}) & 0.643 (\cellcolor{yellow}{↑ 13\%}) & 5.5 \\
        & \Checkmark & \Checkmark & \Checkmark & \textbf{0.327 (\cellcolor{greenl}{↓ 28\%})} & \textbf{0.711 (\cellcolor{greenl}{↑ 25\%})} & 5.5 \\
        \bottomrule
    \end{tabular}
    \caption{Ablation study with two seconds of dedicated training time per object on the \textit{Cube Diorama} dataset. AO denotes Architecture Optimization, ML denotes Meta-learning, and PS denotes Probabilistic ray Sampling. Size represents the disk space serialized NeRF parameters occupy. }
    \label{tab:ablation_study}
\end{table}

We ablate the impact of the three main components of PRENOM by evaluating the upper bound of reconstruction quality using ground-truth object states and camera poses on the \textit{Cube Diorama} dataset. As shown in \Cref{tab:ablation_study}, our per-category architecture optimization achieves a 68\% reduction in model size, averaged over \(\mathcal{I}\), compared to the default tuned category-agnostic architecture adapted from \cite{han2023ro}, while simultaneously improving reconstruction quality. Additionally, meta-learning the parameters and incorporating probabilistic sampling independently contribute further improvements.

We also conduct this ablation with monocular input to highlight the potential benefits of incorporating our priors into monocular reconstruction. 
Given object states, our method enables the transfer of depth-based knowledge acquired during offline training to the online monocular reconstruction process, resulting in even greater improvements than those observed with RGB-D. 
However, accurately estimating object states in monocular setups remains challenging due to scale ambiguity and noisy object association.

\section{Conclusion}
\label{sec_conclusions}

We presented PRENOM, an efficient prior-based neural object mapper that leverages category-level priors to improve 3D object reconstruction and enables canonical pose estimation. By integrating meta-learned priors and probabilistic ray sampling, optimized per-category using a multi-objective genetic algorithm, PRENOM enhances reconstruction accuracy and efficiency, making it well-suited for multi-object mapping in resource-constrained setups. Experimental results demonstrate its effectiveness, achieving a 21\% lower Chamfer distance compared to prior-free obejct-level NeRF-based approaches. Further comparisons with DeepSDF-based approaches show a 13\% improvement in reconstruction metrics, with comparable pose and size estimation capabilities, highlighting the efficiency of our approach, as it was trained for 5$\times$ less time and with a 30\% smaller prior size. However, despite its advantages, PRENOM is inherently limited by its NeRF-based representation, which struggles with out-of-view feature reconstruction. Although our priors help mitigate these issues, the absence of hard constraints to enforce consistency eventually leads to noisier reconstructions, especially if trained for extended periods. Additionally, object state is not refined during training, resulting in slightly worse state estimation than some existing shape prior methods. Future work will focus in these directions.


\bibliographystyle{IEEEtran}
\bibliography{root}

\end{document}